\documentclass{article}
\usepackage{ijcai24}

\usepackage{times}
\usepackage{soul}
\usepackage{url}
\usepackage[hidelinks]{hyperref}
\usepackage[utf8]{inputenc}
\usepackage[small]{caption}
\usepackage{graphicx}
\usepackage{amsmath}
\usepackage{amsthm}
\usepackage{booktabs}
\usepackage{algorithm}
\usepackage{algorithmic}
\usepackage[switch]{lineno}
\usepackage{svg}
\svgsetup{inkscapelatex=false}
\usepackage{tabularx}
\usepackage{multirow}
\usepackage{amsmath}
\usepackage{amssymb}
\usepackage{dsfont}
\usepackage{float}
\usepackage{enumitem}

\usepackage{comment}
\newcommand{\partitle}[1]{\smallskip \noindent \textbf{#1.}}

\title{Contrastive Unlearning: A Contrastive Approach to Machine Unlearning}

\author{
Hong kyu Lee$^1$
\and
Qiuchen Zhang$^2$\and
Carl Yang$^3$\And
Jian Lou$^4$ \And
Li Xiong$^5$\\
\affiliations
$^1$$^2$$^3$$^5$Emory University\\
$^4$Zhejiang University\\
\emails
\{hong.kyu.lee, qiuchen.zhang, j.carlyang\}@emory.edu,
jian.lou@zju.edu.cn,
lxiong@emory.edu
}

\begin{document}

\maketitle

\begin{abstract}

    Machine unlearning aims to eliminate the influence of a subset of training samples (i.e., unlearning samples) from a trained model. Effectively and efficiently removing the unlearning samples without negatively impacting the overall model performance is still challenging. In this paper, we propose a contrastive unlearning framework, leveraging the concept of representation learning for more effective unlearning. It removes the influence of unlearning samples by contrasting their embeddings against the remaining samples so that they are pushed away from their original classes and pulled toward other classes. By directly optimizing the representation space, it effectively removes the influence of unlearning samples while maintaining the representations learned from the remaining samples.
    Experiments on a variety of datasets and models on both class unlearning and sample unlearning showed that contrastive unlearning achieves the best unlearning effects and efficiency with the lowest performance loss compared with the state-of-the-art algorithms.

\end{abstract}

\section{Introduction}

Machine unlearning~\cite{cao_towards_2015} aims to remove a subset of data (i.e., unlearning samples) from a trained machine learning (ML) model and has received increasing attention due to various privacy regulations. Notably, ``the right to be forgotten" from the General Data Protection Requirement (GDPR) gives individuals the right to request their data to be removed from databases, which extends to models trained on such data~\cite{mantelero_eu_2013}. Since models can remember training data within their parameters~\cite{arpit_closer_2017}, it is necessary to ``unlearn" these data from a trained model. 
The goals and evaluation metrics for unlearning typically include: 1) unlearning effectiveness, which measures how well the algorithm removes the influence of unlearning samples. This can be assessed by the model's performance on the unlearning samples (where low accuracy indicates effective unlearning), or by its robustness against membership inference attacks~\cite{shokri_membership_2017}, using unlearning samples (where a low member prediction rate indicates effective unlearning); 2) model performance on its original tasks, which ensures that the unlearning does not significantly degrade its overall accuracy; and 3) computational efficiency, which assesses the time and resources required for the unlearning.

Current machine unlearning approaches can be categorized into \textit{exact unlearning} and \textit{approximate unlearning}. Exact unlearning ensures all influence of the unlearning data is removed as if the data were never part of the training set. Retraining the model from scratch excluding the unlearning samples is a baseline method which can be computationally expensive.  
SISA is an exact unlearning method based on data partitioning and retraining~\cite{bourtoule_machine_2021} which alleviates the computational intensity of complete retraining, but requires training of multiple models (on each partition) and retraining of the models containing the unlearning data, and its partitioning strategy may lead to reduced model performance. Approximate unlearning offers a more feasible alternative and seeks to remove the influence of the unlearning data to a negligible level, typically by updating the model in a way that diminishes the impact of the unlearning data. A subcategory is certified unlearning~\cite{gupta_adaptive_2021,guo_certified_2020,neel_descent--delete_2021} which provides a quantifiable approximation guarantee on the removal of the data. Most approximate unlearning methods use the evaluation metrics discussed earlier to empirically evaluate the models. 

While many promising approaches are proposed, existing works present several limitations: 1) they mainly exploit input and output space and typical classification loss without explicitly considering the latent representations of the samples, 2) they either focus on unlearning samples or remaining samples alone without considering them together or use both but in an ineffective way for unlearning and hence either sacrifice the model performance or the unlearning effectiveness. For example, NegGrad (Negative Gradient) \cite{golatkar_eternal_2020} only uses unlearning samples and attempts to reverse their impact by applying gradient {\em ascent} using the classification loss.
Finetune~\cite{golatkar_eternal_2020} only uses remaining samples to iteratively retrain the model to gradually remove the information of unlearning samples leveraging the catastrophic forgetting effect~\cite{goodfellow_empirical_2013}. SCRUB~\cite{kurmanji_towards_2023} uses both unlearning and remaining samples for unlearning, but requires multiple iterations over the entire remaining samples that leads to excessive computations with poor unlearning effectiveness.

\partitle{Our Contributions}
To address these deficiencies, we present a novel contrastive approach for machine unlearning. We re-purpose the idea of contrastive learning, a widely used representation learning approach, for more effective unlearning.  The main idea is that given an unlearning sample, we contrast it with
1) Positive samples (remaining samples from the same class as the unlearning sample) and push their representations apart from each other, and 2) Negative samples, (remaining samples from different classes as the unlearning sample) and pull their representations close to each other. It has two main insights. First, it exploits the representation space of the samples and directly optimizes the geometric properties of the embeddings of unlearning samples, which captures the underlying structures and most important features of the samples being memorized, facilitating more effective unlearning. 
Second, by contrasting unlearning samples and remaining samples during unlearning and using both positive and negative remaining samples as references for optimizing the embedding of unlearning samples, it can effectively remove the influence of unlearning samples while keeping the embeddings of the remaining samples intact, with an auxiliary classification loss on the contrasted remaining samples, hence maintaining model accuracy. 

Albeit taking inspiration from contrastive learning, our contrastive \emph{unlearning} has novel algorithm designs and gains a new finding, including: 1) we construct contrasting pairs different from conventional contrastive learning to serve the unlearning purpose and further design new contrastive unlearning losses for both sample unlearning and single class unlearning tasks; 2) while it appears common to add a classification loss to maintain the performance of the unlearning model, through the new lens of contrastive unlearning, we make a novel finding that the classification loss can help keep the embeddings of the remaining samples in place and reciprocally improve unlearning effectiveness, which is validated by our empirical analysis followed by in-depth analysis.

\begin{figure}[!h]
    \centering
    \includegraphics[width=\linewidth]{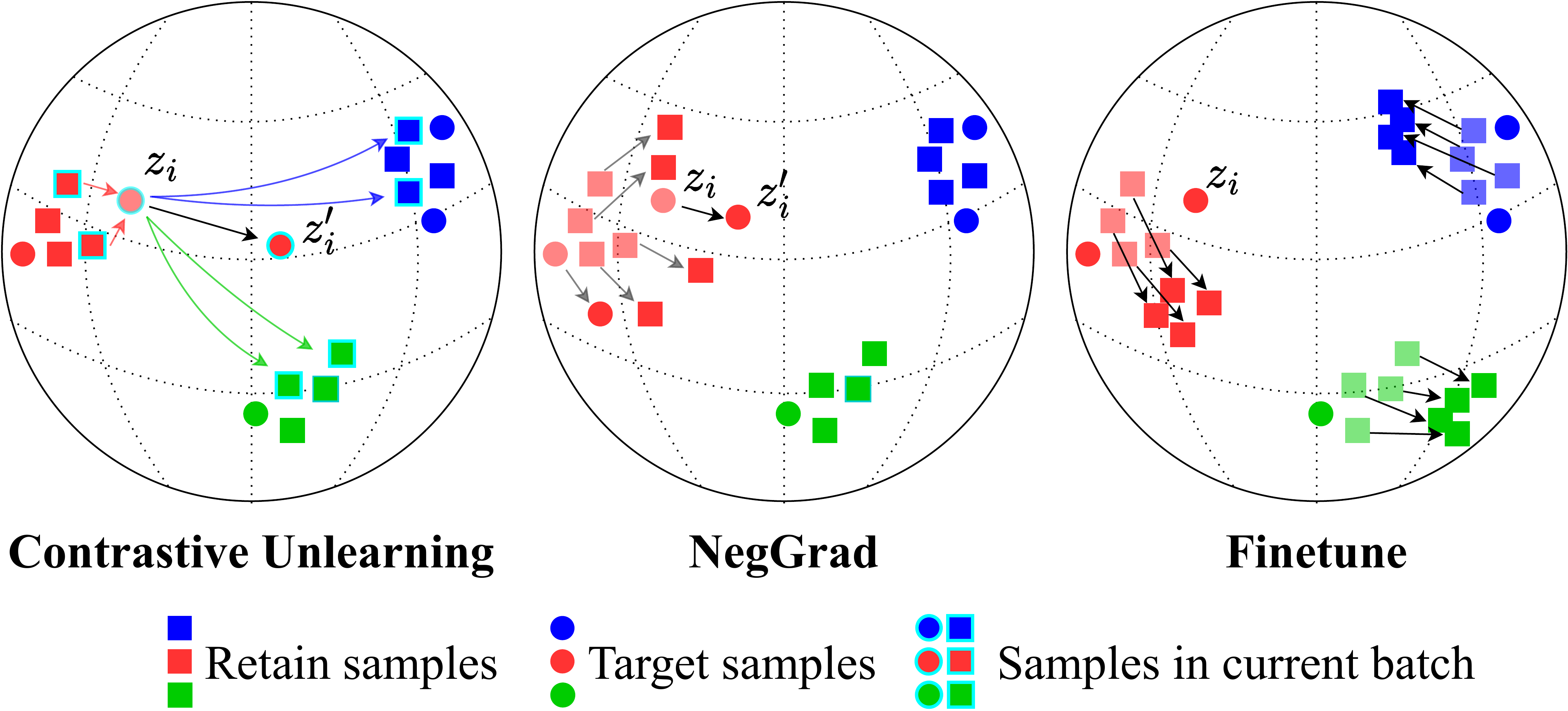}
    \caption{Visualization of representation spaces of contrastive unlearning, gradient ascent, and finetune.}
    \label{fig:csul_01}
\end{figure}

Figure~\ref{fig:csul_01} illustrates the intuition of contrastive unlearning in comparison to existing approaches 
in a normalized representation space. Circles and squares are embeddings of the unlearning samples and remaining samples respectively. The colors represent different classes. We assume the model has been trained, so the embeddings are clustered to their respective classes~\cite{das_separability_2019}. 
Given an embedding of unlearning sample $z_i$, contrastive unlearning pushes $z_i$ away from its own class (positive pairs) and pulls $z_i$ towards the samples with different classes (negative pairs). 
This results in the unlearned embedding $z^\prime_i$ to be geometrically distant from every class (achieving effective unlearning) while keeping the embeddings of remaining samples relatively intact (maintaining model utility).
In comparison, NegGrad attempts to reverse the impact of unlearning samples in the input and output space using classification loss. It has some impact pushing $z_i$ away in the representation space but is not very effective as it changes the decision boundary of classes (ineffective unlearning). In addition, 
it may significantly affect embeddings of remaining samples of the same class (model utility loss). 
Finetune attempts to retrain the model only using remaining samples. In representation space, this only indirectly pushes the unlearning samples away from the remaining samples 
(ineffective unlearning) and is susceptible to overfitting to the remaining samples (model utility loss). 

We conduct comprehensive experiments on both class unlearning and sample unlearning to demonstrate the effectiveness and versatility of our approach in comparison to state-of-the-art methods. Single class unlearning is to remove all samples from a class, and sample unlearning is to remove the arbitrarily selected samples. 
Experimental results on model accuracy show that contrastive unlearning achieves the most effective unlearning (low model accuracy on unlearning samples comparable to the retrained model) while maintaining model utility (high model accuracy on test samples), with high computation efficiency. 
In addition, we conduct a membership inference attack (MIA) \cite{shokri_membership_2017} for deeper verification of unlearning. 
We assume a strong adversary who has full access to the unlearned model, simulating an administrator who conducted unlearning and wants to verify the effectiveness of unlearning~\cite{thudi_necessity_2022,cotogni_duck_2023}. 
 Contrastive unlearning has the lowest member prediction rate on unlearning samples. compared to all baselines, indicating the most effective unlearning.

In summary, our contributions are as follows.

\begin{enumerate}[leftmargin=*]
    \item We propose contrastive unlearning, an unlearning algorithm utilizing the concept of contrastive learning. It directly optimizes the geometric properties of embeddings of unlearning samples by contrasting them with embeddings of the remaining samples in the representation space. This effectively captures and removes the most important features relevant for classification from the embeddings of the unlearning samples (achieving effective unlearning) while keeping the embeddings of remaining samples relatively intact (maintaining model utility). In addition, we design contrastive learning losses for both single class unlearning and sample unlearning. 
    
    \item We conduct comprehensive experiments comparing contrastive unlearning with various state-of-the-art methods on two unlearning tasks, single class and sample unlearning, to demonstrate the effectiveness and versatility of our approach. We also conduct a membership inference attackto verify the effectiveness of unlearning. The results show that contrastive unlearning is most effective in unlearning while maintaining model utility with high computation efficiency. 
    
    
    
\end{enumerate}

\section{Related Works}

Machine unlearning is introduced by~\cite{cao_towards_2015}. They defined two unlearning goals: completeness suggests that an unlearning algorithm should reverse the influence of unlearning samples and the unlearned model should be consistent with a model retrained from all training samples except the unlearning samples; timeliness requires the running time of the unlearning algorithm should be faster than retraining. On top of that, the unlearned model should incur low performance loss due to the unlearning.

Exact unlearning ensures the completeness of unlearning. 
SISA is an exact unlearning framework based on sharding. It splits the dataset into multiple partitions and trains a model for each shard. Given an unlearning request, it retrains the models whose shard has the unlearning sample~\cite{bourtoule_machine_2021}. ARCANE uses a partitioning strategy by the classes of the samples~\cite{yan_arcane_2022}. These frameworks require partitioned training and still expensive retraining computation, and model performance is highly dependent on partitioning strategy~\cite{koch_no_2023}.

Approximate unlearning allows approximate completeness. 
Certified unlearning provides a mathematical guarantee on the approximation. $(\varepsilon, \delta)$-indistinguishability similar to differential privacy is proposed~\cite{guo_certified_2020} using Newton-type hessian update. 
~\cite{neel_descent--delete_2021} proposes an algorithm based on project gradient descent on the partitioned dataset with a probabilistic bound. Approximation guarantee is also useful for graph unlearning~\cite{wu_certified_2023}. \cite{gupta_adaptive_2021} studied unlearning requests that may be correlated and derived the unlearning guarantee with adaptive unlearning streams. Fisher unlearning uses Fisher information matrix~\cite{golatkar_eternal_2020} to identify optimal noise to remove the influence of unlearning samples. A general drawback of certified unlearning algorithms is the difficulty to scale to neural networks, as convexity of the loss function is often required to satisfy the mathematical guarantee. Also, they are computationally expensive. Despite some efforts such as LCODEC~\cite{mehta_deep_2022} for alleviating the computation cost by selectively generating Hessians, computing Fisher information and Hessian matrix is expensive. 

Another body of approximate unlearning shows the unlearning effect through empirical evaluations. Besides NegGrad \cite{golatkar_eternal_2020} and 
Finetune~\cite{golatkar_eternal_2020} discussed earlier, several methods are designed to unlearn an entire class. UNSIR~\cite{tarun_fast_2023} conducts noisy gradient updates using the unlearning class. 
Boundary unlearning unlearns an entire class~\cite{chen_boundary_2023} by changing decision boundaries. ERM-KTP uses a special neural architecture known as entanglement reduce mask~\cite{lin_erm-ktp_2023}. 
SCRUB~\cite{kurmanji_towards_2023} is based on the teacher-student network in which the teacher or the original model transfers knowledge to the unlearned model in every class except the unlearning class. 

Our approach is an approximate unlearning method that works for both sample and class unlearning. We compare it with both types of methods, as well as empirical and certified methods, and demonstrate its superiority through empirical evaluations. 

\section{Preliminaries and Problem Definition}

\subsection{Contrastive Learning}

Contrastive learning, in particular SimCLR, pulls an embedding of a sample (or an anchor) toward the embedding of its augmented self and pushes it away from embeddings of other samples~\cite{chen_simple_2020}. SimCLR was originally proposed as a self-supervised learning framework, hence it does not rely on labels. To enhance classification performance, Supervised Contrastive learning (SupCon) was introduced to leverage contrasting learning in a supervised manner~\cite{khosla_supervised_2020}. Instead of contrasting based on augmented data, it contrasts samples based on their classes. Specifically, embeddings of the same classes are pulled together, and embeddings of different classes are pushed away. We utilize a contrastive loss inspired by SupCon for unlearning. 


\subsection{Problem Definition}

We define a classification model $\mathcal{F} = \mathcal{H}\left(\mathcal{E_\theta\left(\cdot\right)}\right)$ where $\mathcal{E}_\theta\left(\cdot\right)$ is a neural network based encoder parameterized by $\theta$ and $\mathcal{H}\left(\cdot\right)$ is a classification head. $\mathcal{E}_\theta$ produces embeddings $z$ given a sample $x$. $\mathcal{H}$ receives $z$ and yields a prediction. Let $\mathcal{F}$ be trained using dataset $\mathcal{D}_{tr} = \{\left(x_1, y_1\right) \cdots \left(x_n, y_n\right)\}$, where each data point is a tuple $\left(x_i, y_i\right)$ including feature set $x_i$ and  label $y_i \in \{0 \cdots C\}$ where $C$ is the number of classes. We suppose $\mathcal{F}$ was trained with cross-entropy loss. Let $\mathcal{D}_{ts}$ be a test dataset sampled from an analogous distribution with $\mathcal{D}_{tr}$, satisfying $\mathcal{D}_{ts} \cup \mathcal{D}_{tr} = \emptyset$.

Let $\mathcal{D}^u_{tr}$ $\subseteq \mathcal{D}_{tr}$ be a set of samples to be forgotten (i.e., unlearning samples). The remaining set is $\mathcal{D}^r_{tr} = \mathcal{D}_{tr} \setminus \mathcal{D}^u_{tr}$. Let a retrained model $\mathcal{F}^R$ be trained only with $\mathcal{D}^r_{tr}$. An unlearning algorithm $M$ receives $\mathcal{D}^r_{tr}, \mathcal{D}^u_{tr}, \theta$ and produces $\theta^\prime$. An unlearned model $\mathcal{F}^\prime = \mathcal{H}\left(\mathcal{E}_{\theta^\prime}\right)$ should resemble $\mathcal{F}^R$. 

\partitle{Single Class Unlearning}
For single class unlearning, $D^u_{tr}$ consists of entire samples of an unlearning class $c$. 
The test set $\mathcal{D}_{ts}$ can be split into $\mathcal{D}^u_{ts}$ and $\mathcal{D}^r_{ts}$, where $\mathcal{D}^u_{ts}$ includes all test samples of class $c$, and $\mathcal{D}^r_{ts} = \mathcal{D}_{ts} \setminus \mathcal{D}^u_{ts}$ includes all test samples of remaining classes. 
A retrained model $\mathcal{F}^R$ will have zero accuracy on $\mathcal{D}^u_{tr}$ and $\mathcal{D}^u_{ts}$, the training and test samples of class $c$, since it was retrained without class $c$. 
So given an accuracy function $Acc$, the goal of single class unlearning is for the unlearned model $\mathcal{F}^\prime$ to achieve accuracy as close to zero as possible on both training and test samples of class $c$ (effective unlearning) and similar accuracy as the retrained model $\mathcal{F}^R$ for remaining classes (model performance).  
 \begin{small}
     \begin{flalign}
        & \mathrm{Acc}\left(\mathcal{F}^\prime, \mathcal{D}^u_{tr}\right) \approx 0, \quad \mathrm{Acc}\left(\mathcal{F}^\prime, \mathcal{D}^u_{ts}\right) \approx 0, \\
        & \mathrm{Acc}\left(\mathcal{F}^\prime, \mathcal{D}^r_{ts}\right) \approx Acc\left(\mathcal{F}^R, \mathcal{D}^r_{ts}\right).
         \label{def:scu}
    \end{flalign}
 \end{small}

\partitle{Sample Unlearning}
For sample unlearning, the unlearning samples $\mathcal{D}^u_{tr}$ can belong to different classes. 
A retrained model $\mathcal{F}^R$ will have similar accuracy on unlearning samples $\mathcal{D}^u_{tr}$ and test samples $\mathcal{D}_{ts}$ since unlearning samples are not in the training set anymore. So the goal of sample unlearning is for the unlearned model $\mathcal{F}^\prime$ to achieve similar accuracy as the retrained model $\mathcal{F}^R$ on both unlearning samples (effective unlearning) and test samples (model performance).  
 \begin{small}
     \begin{flalign}
        & \mathrm{Acc} \left(\mathcal{F}^\prime, \mathcal{D}^u_{tr}\right) \approx 
        Acc\left(\mathcal{F}^R, \mathcal{D}_{ts}\right),
        \\
        & \mathrm{Acc} \left(\mathcal{F}^\prime, \mathcal{D}_{ts}\right) \approx Acc\left(\mathcal{F}^R, \mathcal{D}_{ts}\right).
    \end{flalign}
 \end{small}
\section{Contrastive Unlearning}
The novelty of contrastive unlearning is our perspective on utilizing geometric properties of latent representation space for unlearning purposes. If a sample $x$ had been used as a training example, information extracted from $x$ by $\mathcal{E}_\theta$ would be expressed as geometric properties in the representation space. Specifically, we hypothesize that a trained model generates geometrically similar embeddings from samples with the same class and distant embeddings for samples with different classes even when the model was not trained by representation learning techniques. This can be supported by existing literature, which mathematically and empirically showed that a model optimized with cross-entropy loss produces higher geometric similarity among embeddings of samples of the same class and lower similarity among different classes~\cite{das_separability_2019,graf_dissecting_2021}. 



From this intuition, we aim to modify characteristics of representation space by pushing embeddings of unlearning samples away from those of remaining samples. To effectively achieve this, we contrast each unlearning sample with 1) remaining samples from the same class (positive pairs) and push their representations apart from each other,
and 2) remaining samples from different classes (negative pairs) and pull their representations close to each
other. 
To this end, the embeddings of unlearning samples end up in the middle of all the remaining samples. This has some relation with existing literature of contrastive learning, however, our approach is fundamentally different as it contrasts pairs of unlearning and remaining samples while contrastive learning contrasts samples simply by their classes. 


\partitle {Contrastive Unlearning Loss: Sample Unlearning}
Contrastive unlearning uses a batched process. In each round, an unlearning batch $X^u = \{x^u_1, \cdots x^u_B\}$ with size $B$ is sampled from the unlearning data $\mathcal{D}^u_{tr}$, and a remaining batch $X^r = \{x^r_1 \cdots x^r_B\}$ is sampled from the remaining set $\mathcal{D}^r_{tr}$. We denote $x_i$ as $i$-th sample of $X^u$ as an anchor. Based on the anchor $x_i$, positives and negatives are chosen from $X^r$. Positives are $P_\mathbf{x}\left(x_i\right) = \{x_j \vert x_j \in X^r, y_j = y_i \}$, or remaining samples with the same class as $x_i$; negatives are $N_\mathbf{x}\left(x_i\right) = \{ x_j \vert x_j \in X^r,  y_j \neq y_i\}$, or remaining samples with different class as $x_i$. Correspondingly, let embeddings of positives and negatives be $P_\mathbf{z}\left(x_i\right) = \{z_j \vert z_j=\mathcal{E}_{\theta}\left(x_j\right), x_j \in P_\mathbf{x}\left(x_i\right)\}$ and $N_\mathbf{z}\left(x_i\right) = \{z_j \vert z_j=\mathcal{E}_{\theta}\left(x_j\right), x_j \in N_\mathbf{x}\left(x_i\right)\}$. The contrastive unlearning loss aims to minimize the similarity of positive pairs and maximizes the similarity of negative pairs (the opposite of contrastive learning). 
\begin{small}
    \begin{flalign}
    \mathcal{L}_{UL} &= \sum_{x_i \in X^u}{\frac{-1}{\lvert N_\mathbf{z}\left(x_i\right) \rvert}} \sum_{z_a \in N_z}{\log{\frac{\mathrm{exp}\left(z_i \cdot z_a / \tau \right)}{\sum\limits_{z_p \in P_\mathbf{z}\left(x_i\right)}{\mathrm{exp}\left(z_i \cdot z_p / \tau \right)}}}}
    \label{eq:cul}
\end{flalign}
\end{small}
\noindent where $\tau \in \mathcal{R}^+$ is a scalar temperature parameter.

\partitle{Contrastive Unlearning Loss: Single Class Unlearning}
For single class unlearning, 
the unlearning set $\mathcal{D}^u_{tr} = \{\left(x_i, y_i\right)\vert y_i = c\}$ and remaining set $\mathcal{D}^r_{tr} = \{\left(x_i, y_i\right)\vert y_i \neq c\}$. This makes the positive set $P_\mathbf{z} = \emptyset$ as none of remaining samples belong to class $c$. In short, there are no positive remaining samples to push away the unlearning samples. Thus we change equation~\ref{eq:cul} as follows.
\begin{small}
\begin{flalign}
    \mathcal{L}_{UL} &= \sum_{x_i \in X^u}{\frac{-1}{\lvert N_\mathbf{z}\left(x_i\right) \rvert}} \sum_{z_a \in N_z}{\log{\frac{\mathrm{exp}\left(z_i \cdot z_a / \tau \right)}{\lvert N_z\left(x_i\right) \rvert}}}.
    \label{eq:cul_2}
\end{flalign}
\end{small}
We replaced the previous denominator to $\lvert N_\mathbf{z}\left(x_i\right) \rvert$. This is because equation~\ref{eq:cul} requires both directions to push and pull unlearning samples. Lacking one of the directions increases the instability of the loss. Since $P_z = \emptyset$, we replace the denominator to $\lvert N_\mathbf{z}\left(x_i\right)\rvert$ to introduce damping effects against excessively pulling unlearning samples to negative samples.

\partitle{Classification Loss of Remaining Samples} A novel challenge of contrastive unlearning is to preserve embeddings of remaining samples. Optimizing equation~\ref{eq:cul} not only alters embeddings of the anchor unlearning sample but also reciprocally alters embeddings of all samples in $P_{\mathbf{x}}$ and $N_\mathbf{x}$. All positive samples are slightly pushed away from and all negatives are slightly pulled toward the anchor. A similar effect arises in contrastive learning, but it is not problematic as it reinforces the consolidation of embeddings of the same class, which is a desired effect. However, for unlearning purposes, embeddings of $X^r$ have to be preserved, because: 1) not preserving them directly leads to a loss in model performance, and 2) it also reciprocally affects unlearning effectiveness as magnitude of pulling and pushing decreases. In short, embeddings of $X^r$ are also modified as a byproduct of optimization and it is necessary to restore them back. We utilize cross-entropy loss for restoring embeddings of $X^r$, because it derives maximum likelihood independently to each sample~\cite{shore_properties_1981}. This ensures obtaining directions very close to the original embeddings no matter how embeddings of remaining samples are modified. Combining the unlearning loss, the final loss for our proposed contrastive unlearning is as follows,
\begin{small}
    \begin{flalign}
    \mathcal{L} = \lambda_{UL} \mathcal{L}_{UL} + \lambda_{CE} \mathcal{L}_{CE}\left(\mathcal{F}\left(X^r\right), Y^r\right),
\end{flalign}
\end{small}
\noindent where $X^r$ and $Y^r$ are batched remaining samples and their corresponding labels. $\lambda_{CE}$ and $\lambda_{UL}$ are hyperparamters to determine influence of two loss terms.  


\begin{algorithm}[tb]
\scriptsize
    \caption{Contrastive Unlearning}
    \label{alg:algo_01}
    \textbf{Input}: $\theta, \mathcal{H}\left(\cdot\right), \mathcal{E}\left(\cdot\right), D^r_{tr}, D^u_{tr}$, $\mathcal{D}_{\mathrm{eval}}$\\
    \textbf{Parameter}: $iter, \lambda_{CL}, \lambda_{UL}, \omega$\\
    \textbf{Output}: $\theta^\prime$
    \begin{algorithmic}[1] 
        \WHILE{termination condition is not satisfied}
            \FOR{ each $X^u \in D^u_{tr}$}
                \FOR {$1, \cdots, \omega$}
                    \STATE Sample $\left(X^r, Y^r\right)$ from $\mathcal{D}^r_{tr}$
                    \STATE Determine $P_\mathbf{z}\left(x_i\right), N_\mathbf{z}\left(x_i\right)$ $\forall x_i \in X^u$
                    \STATE $\ell_{CE} \leftarrow \mathcal{L}_{CE}\left(\mathcal{H}\left(\mathcal{E}_\theta\left(X^r\right)\right), Y^r\right)$
                    \STATE $\ell_{UL} \leftarrow \lambda_{UL} \mathcal{L}_{UL}\left(P_\mathbf{z}\left(x_i\right), N_\mathbf{z}\left(x_i\right)\right)$ $\forall x_i \in X^u$
                    \STATE $\theta \leftarrow \theta - \eta\nabla\left(\ell_{CE} + \ell_{UL}\right)$
                \ENDFOR
            \ENDFOR
            \STATE $\theta^\prime \leftarrow \theta$
            \STATE Evaluate, get termination condition $\theta^\prime$ with $\mathcal{D}_{\mathrm{eval}}$
        \ENDWHILE

        \STATE \textbf{return} $\theta^\prime$
    \end{algorithmic}
\end{algorithm}

\partitle{Complete Algorithm} Algorithm~\ref{alg:algo_01} shows step-wise overview of contrastive unlearning. It iterates for all unlearning batches $X^u$ in $D^u_{tr}$. For each $X^u$, it computes unlearning loss by sampling a random remaining batch $X^r$ for contrasting purposes. For each $X^u$, sampling and loss derivation are repeated $\omega$ times. Higher $\omega$ stabilizes the unlearning procedure by contrasting unlearning samples against multiple sets of remaining samples. From the experiment, we set $\omega$ to be at most 4 to reduce computational overhead and our algorithm showed stable unlearning performance. 

\partitle{Termination Condition}
The termination condition for the algorithm differs based on the task of unlearning. We assume a small dataset $\mathcal{D}_{\mathrm{eval}}$ is available for evaluation. The algorithm evaluates $\mathcal{F}^\prime$ with $\mathcal{D}_{eval}$ and terminates if it satisfies unlearning criteria. For single class unlearning, $\mathcal{D}_{\mathrm{eval}} = D^u_{ts}$, the test data of the unlearning class. The algorithm terminates when the accuracy of 
the unlearned model $\mathcal{F}^\prime$ on the unlearning class falls below a threshold  where $C$ is the total number of classes in the training data and 1/$C$ corresponds to the accuracy of a random guess.
\begin{small}
    \begin{flalign}
    \mathrm{Acc}\left(\mathcal{F}^\prime, \mathcal{D}_{\mathrm{eval}} \right) \leq \frac{1}{C}.
\end{flalign}
\end{small}
For sample unlearning, $\mathcal{D}_{\mathrm{eval}} = \{\mathcal{D}^u_{eval}, \mathcal{D}^{ts}_{eval} \}$ where $\mathcal{D}^u_{eval} \subseteq \mathcal{D}^u_{tr}$ and $ \mathcal{D}^{ts}_{eval} \subseteq \mathcal{D}_{ts}$. The algorithm terminates when  the accuracy of $\mathcal{F}^\prime$ on the unlearning samples $\mathcal{D}^u_{eval}$ drops below the accuray on test samples $\mathcal{D}^{ts}_{eval}$. 
\begin{small}
    \begin{flalign}
    \mathrm{Acc}\left(\mathcal{F}^\prime, \mathcal{D}^u_{eval}\right) 
    \leq 
    \mathrm{Acc}\left(\mathcal{F}^\prime, \mathcal{D}^{ts}_{eval} \right).
    \label{eq:random_sample_stop_cond}
\end{flalign}
\end{small}
It is not desired to terminate the algorithm before satisfying this condition because it implies that the model still retains information regarding $\mathcal{D}^u_{tr}$. It is also not desired to continue running the algorithm to further reduce accuracy on $\mathcal{D}^u_{tr}$ much lower than $\mathcal{D}_{ts}$ because it is negatively injecting information regarding $\mathcal{D}^u_{tr}$ into $\theta^\prime$. This results in $\mathcal{F}^\prime$ to deliberately make incorrect classification on $\mathcal{D}^u_{ts}$, which is not aligned with the goal of sample unlearning. 

\section{Experiments}

\subsection{Experiment Setup}
\partitle{Datasets and Models} We use two standard benchmark datasets, CIFAR-10 and SVHN, and use ResNet(RN)-18, 34, 50, and 101 models~\cite{he_deep_2016} in our experiments. We train each model with each dataset without any data augmentation except normalization. Readers may refer to the appendix for the performance of original models. 

\partitle{Comparison Methods} 
For class unlearning, we remove all samples belong to class 5 and for sample unlearning, we remove randomly selected 500 samples. For both class unlearning and sample unlearning tasks, we use \textbf{Retrain}, a retrained model using the training data excluding the unlearning class or samples, as an ideal reference for unlearning completeness and model performance. 
We include four state-of-the-art methods designed for \textbf{sample unlearning}: 1) \textbf{Finetune}~\cite{golatkar_eternal_2020}
       leverages catastrophic forgetting~\cite{goodfellow_empirical_2013} and iteratively trains the original model only using the remaining samples. 
2) \textbf{Neggrad}~\cite{golatkar_eternal_2020}
        conducts gradient ascent using unlearning samples.
3) \textbf{Fisher}~\cite{golatkar_eternal_2020}
         is a certified unlearning algorithm using randomization techniques borrowed from differential privacy and  leverages the Fisher information matrix to design optimal noise for noisy gradient updates.
4)  \textbf{LCODEC}~\cite{mehta_deep_2022}
        is also a certified unlearning method that proposes a fast and effective way of obtaining Hessian by selecting parameters by their importance.

We include four state-of-the-art methods specifically designed for \textbf{single class unlearning}: 
1) \textbf{Boundary Expansion}~\cite{chen_boundary_2023}
        trains the model using all unlearning samples as a temporary class and then discards the temporary class. 
2) \textbf{Boundary Shrink}~\cite{chen_boundary_2023}
        is similar to Boundary Expansion but it modifies the decision boundary of unlearning class to prevent unlearning samples from being classified into the unlearning class (unlearning samples are classified as other classes).     3) \textbf{SCRUB}~\cite{kurmanji_towards_2023}
        is based on the teacher-student framework and  
        selectively transfers information from the original model to the unlearned model (all information except that of the unlearning class).  
   4) \textbf{UNSIR}~\cite{tarun_fast_2023}
        uses an iterative process of impairing and recovering and generates noise that maximizes error in the unlearning class and repairs the classification performance for the other classes.

We note that sample unlearning algorithms may be used for class unlearning. However, the class unlearning baselines we have chose here already demonstrated their superiority over the sample unlearning methods including Finetune, Neggrad, and Fisher, hence we do not include them in comparison.  

\partitle{Evaluation Metrics} We evaluate model performance, unlearning effectiveness, and efficiency of the algorithms. 
\begin{itemize}[leftmargin=*]
    \item \textbf{Model performance} is assessed by accuracy of the unlearned model on the test data of remaining classes $\mathcal{D}^r_{ts}$ (class unlearning) and on the test data $\mathcal{D}_{ts}$ (sample unlearning). The accuracy should be similar to the retrained model.
    \item \textbf{Unlearning effectiveness} is assessed by accuracy of the unlearned model on the training and test data of unlearning class $\mathcal{D}^u_{tr}$ and $\mathcal{D}^u_{ts}$ (class unlearning) and the unlearning samples $\mathcal{D}^u_{tr}$ (sample unlearning). The lower the accuracy, the more effective the unlearning. We also conduct MIA for further unlearning verification to be described next. 
    \item \textbf{Efficiency} is measured by the  
    runtime of the unlearning algorithm. A shorter runtime indicates better efficiency.
\end{itemize}

\partitle{Unlearning Verification via MIA}
We conduct a membership inference attack (MIA)~\cite{shokri_membership_2017} to verify sample unlearning. 
We assume an adversary with full access to the unlearned model and training data, simulating an administrator who conducted unlearning and uses MIA to verify the effectiveness of unlearning~\cite{thudi_necessity_2022,cotogni_duck_2023}.

To train the attack model, we sample  $\mathcal{D}^M$ from remaining samples $\mathcal{D}^r_{tr}$ (as members) and $\mathcal{D}^N$ from test samples $\mathcal{D}_{ts}$ (as non-members). An attack model is trained with both members and non-members using their output from the unlearned model $\{\mathcal{F}^\prime\left(\mathbf{x}\right) \vert \mathbf{x} \in \mathcal{D}^M \cup \mathcal{D}^N\}$ as features and labels as $\{\mathbf{y}_i\vert\mathbf{y}_i=1 \; \forall x_i\in \mathcal{D}^M, \mathbf{y}_i=0 \; \forall {x}_i \in \mathcal{D}^N \}$. We then test the attack model on the unlearning samples $\mathcal{D}^u_{tr}$ and selected test member samples from remaining samples $\mathcal{D}^r_{tr}$.   
We report the \textbf{Member prediction rate} defined as number of positive (member) predictions by the MIA divided by total number of tests. 
 It can be considered as false positive rate (FPR) for unlearning samples (considering them as non-members) and true positive rate (TPR) for members. 
An effective unlearning algorithm should have a low member prediction rate on unlearning samples and  high member prediction rate on member samples. 
Our metric is consistent with existing literature \cite{jia2023model} utilizing true negative rate (TNR) for unlearning samples and test non-member samples (considering both as non-members), which essentially measures the opposite to ours, i.e., considering non-members rather than members. We focus on predicting the members because MIA is designed to infer members.


\subsection{Results on Single Class Unlearning}

\begin{table}
\scriptsize
    \centering
    \setlength{\tabcolsep}{1pt}
    \begin{tabularx}{0.5\textwidth}{*{8}{>{\centering\arraybackslash}X}}
        \toprule
        \scalebox{0.9}{Model} & \scalebox{0.9}{Evaluation} & \scalebox{0.9}{\begin{tabular}{c}Retrain\\ (reference) \end{tabular}} & \scalebox{0.9}{\textbf{Contrastive}} & \scalebox{0.9}{\begin{tabular}{c}Boundary\\ Shrink \end{tabular}} & \scalebox{0.9}{\begin{tabular}{c}Boundary\\Expansion \end{tabular}} & \scalebox{0.9}{SCRUB} & \scalebox{0.9}{UNSIR} \\
        \midrule 
        \multirow{3}{*}{RN18} & $\mathcal{D}^r_{ts}$  & 86.96 &  85.79 & 83.62 & 82.34 & 83.91 & 57.36 \\
                                   &  $\mathcal{D}^u_{tr}$ & 0.00  &  0.00  & 4.54  & 0.00  & 35.42 & 0.00\\
                                   &  $\mathcal{D}^u_{ts}$ & 0.00  &  0.00  & 4.62  & 6.51 & 9.30 & 0.00\\
        \midrule
        \multirow{3}{*}{RN34} & $\mathcal{D}^r_{ts}$  & 88.01 & 86.59 & 84.70 & 83.19 & 82.22 & 47.02 \\
                                   &  $\mathcal{D}^u_{tr}$ & 0.00 & 0.00  & 2.46 & 0.00& 3.18 & 0.00\\
                                   &  $\mathcal{D}^u_{ts}$ & 0.00 & 0.00  & 4.60 & 6.81 & 0.80& 0.00\\
        \midrule
        \multirow{3}{*}{RN50} & $\mathcal{D}^r_{ts}$  & 87.78 & 87.98 & 85.52 & 83.39 & 84.44 & 37.41 \\
                                   &  $\mathcal{D}^u_{tr}$ & 0.00& 0.00& 2.74 & 0.00& 7.16 & 0.00\\
                                   &  $\mathcal{D}^u_{ts}$ & 0.00& 0.00& 5.90 & 8.22 & 1.51 & 0.00\\
        \midrule
        \multirow{3}{*}{RN101} & $\mathcal{D}^r_{ts}$  & 87.94 & 88.69 & 83.91 & 82.48 & 85.03 & 42.40 \\
                                   &  $\mathcal{D}^u_{tr}$ & 0.00   & 0.00  &  4.91 &   0.00& 13.46 & 0.00\\
                                   &  $\mathcal{D}^u_{ts}$ & 0.00   & 0.00  &  7.25 &  8.50 &  4.55 & 0.00\\    
        \bottomrule
    \end{tabularx}
    \caption{Performance evaluation for single class unlearning on CIFAR-10.}
    \label{tab:scl_cifar10}
\end{table}

\begin{table}
\scriptsize
    \centering
    \setlength{\tabcolsep}{1pt}
        \begin{tabularx}{0.5\textwidth}{*{19}{>{\centering\arraybackslash}X}}
        \toprule
        \scalebox{0.9}{Model} & \scalebox{0.9}{Evaluation} & \scalebox{0.9}{\begin{tabular}{c}Retrain\\ (reference) \end{tabular}} & \scalebox{0.9}{\textbf{Contrastive}} & \scalebox{0.9}{\begin{tabular}{c}Boundary\\ Shrink \end{tabular}} & \scalebox{0.9}{\begin{tabular}{c}Boundary\\Expansion \end{tabular}} & \scalebox{0.9}{SCRUB} & \scalebox{0.9}{UNSIR} \\
    
        \midrule 
        \multirow{3}{*}{RN18} & $\mathcal{D}^r_{ts}$  & 95.43 & 93.91 & 94.84 & 93.71 & 93.88 & 90.3 \\
                                   &  $\mathcal{D}^u_{tr}$ & 0.00& 0.00& 29.79 & 80.25 & 88.67 & 0.00\\
                                   &  $\mathcal{D}^u_{ts}$ & 0.00& 0.00& 37.46 & 2.61 & 77.39 & 0.00\\
        \midrule
        \multirow{3}{*}{RN34} & $\mathcal{D}^r_{ts}$  & 95.46 & 94.33 & 95.12 & 94.50 & 94.57 & 85.82 \\
                                   &  $\mathcal{D}^u_{tr}$ & 0.00& 0.00   & 34.69 & 63.92 & 0.96 & 0.00\\
                                   &  $\mathcal{D}^u_{ts}$ & 0.00& 0.00   & 41.99 & 4.27 & 0.42 & 0.00\\
        \midrule
        \multirow{3}{*}{RN50} & $\mathcal{D}^r_{ts}$  & 95.83 & 94.87 & 95.47 & 95.01 & 93.75 & 70.56 \\
                                   &  $\mathcal{D}^u_{tr}$ & 0.00  & 0.00  & 40.01 & 3.92  & 2.68 & 0.00\\
                                   &  $\mathcal{D}^u_{ts}$ & 0.00  & 0.00  & 42.37 & 8.74  & 9.64 & 0.00\\
        \midrule
        \multirow{3}{*}{RN101} & $\mathcal{D}^r_{ts}$  & 96.16 & 94.90 & 95.65 & 95.07 & 94.65 & 83.90 \\
                                   &  $\mathcal{D}^u_{tr}$ & 0.00   &  0.00 & 42.77 & 51.53 & 0.00& 0.00\\
                                   &  $\mathcal{D}^u_{ts}$ & 0.00   &  0.00 & 45.39 & 3.94  & 0.00& 0.00\\
                        
        \bottomrule
    \end{tabularx}
    \caption{Performance evaluation for single class unlearning on SVHN.}
    \label{tab:scl_svhn}
\end{table}

\partitle{Unlearning Effectiveness and Model Performance}
Table~\ref{tab:scl_cifar10} depicts accuracy of different unlearned models on $\mathcal{D}^r_{ts}$ (test set of remaining classes), $\mathcal{D}^u_{tr}$ (train set of unlearning class), and $\mathcal{D}^u_{ts}$ (test set of unlearning class) on CIFAR-10 for a randomly selected class 5. We experimented with all classes and they show similar performances. 
The retrain model shows the expected results with stable accuracy on $\mathcal{D}^r_{ts}$ (similar to the accuracy of original models shown in the Appendix)  and zero for both $\mathcal{D}^u_{tr}$ and $\mathcal{D}^u_{ts}$ since the class has been removed from training. 
Among all methods, contrastive unlearning is the only one that achieves 0 accuracy on the unlearning class indicating complete unlearning while preserving the accuracy on the remained classes. UNSIR is the only baseline achieving 0 accuracy in the unlearning class, however, it suffers from a significant performance loss. All other methods fail to completely remove the influence while also showing a performance loss in the remaining classes.  

Table~\ref{tab:scl_svhn} illustrates accuracy of unlearned models on SVHN dataset. It shows a similar trend as the CIFAR-10 dataset. 
UNSIR provides better performance on the SVHN dataset because features of SVHN are easier to learn thus the model suffers less utility loss than CIFAR-10. However, it still suffers a significantly higher utility loss than contrastive unlearning. All other baselines show a high accuracy on the unlearning class in many cases, indicating they failed to remove the influence of the unlearning class.  Contrastive unlearning consistently removed all influence of unlearning class with a negligibly small loss of performance.

\begin{figure}
    \centering
    \includegraphics[width=0.7\linewidth]{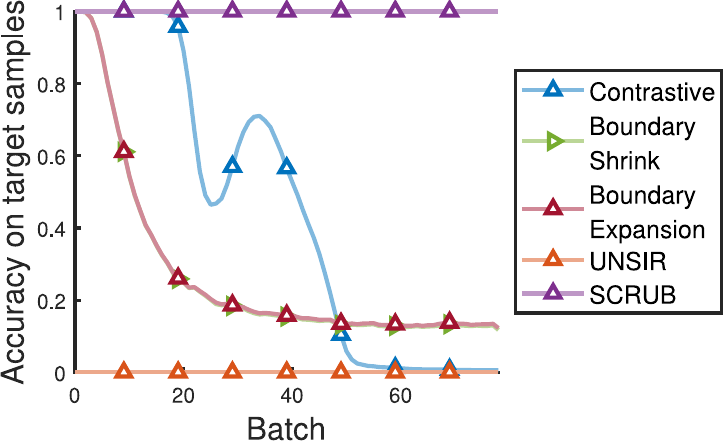}
    \caption{Accuracy on unlearning class vs.  number of batches on $\mathcal{D}^u_{tr}$.}
    \label{fig:sc_unlearn}
\end{figure}

\begin{table}
    \centering
    \scriptsize
    \setlength{\tabcolsep}{1pt}
    \begin{tabularx}{0.5\textwidth}{*{7}{>{\centering\arraybackslash}X}}
        \toprule
        Model & Retrain & \textbf{Contrastive} & \begin{tabular}{c}Boundary\\ Shrink \end{tabular} & \begin{tabular}{c}Boundary\\Expansion \end{tabular} & SCRUB & UNSIR \\
        \midrule
        RN18 & 1566.36 & \textbf{48.90} & 105.22 & 112.87 & 150.40 & 59.98 \\
        RN34 & 2072.76 & \textbf{75.45} & 181.12 & 139.90 & 240.39 & 90.58 \\
        RN50 & 3820.62 & \textbf{105.41} & 315.69 & 240.44 & 435.49 & 169.89 \\
        RN101 & 7493.79 & \textbf{139.94} & 540.21 & 425.77 & 747.65 & 270.38 \\
        \bottomrule
    \end{tabularx}
    \caption{Processing time of class unlearning algorithms on CIFAR-10 dataset (in seconds).}
    \label{tab:sc_time}
\end{table}

\partitle{Efficiency} Figure~\ref{fig:sc_unlearn} shows the progress of the unlearning algorithms in terms of the accuracy on unlearning class $\mathcal{D}^u_{tr}$ vs. the number of batches in a single epoch. Both contrastive unlearning and other baselines are designed to run unlearning procedures multiple times for each batch. However, we fixed the hyperparameters of each algorithm so that each batch of $\mathcal{D}^u_{tr}$ is processed only once. Reaching faster to zero accuracy indicates that the algorithm is more efficient, as it needs a smaller number of batches to achieve unlearning.
The figure shows that contrastive unlearning reaches zero approximately at the 60th batch while boundary shrink and boundary expansion still show approximately 10\% accuracy after the first epoch. UNSIR shows zero accuracy from the beginning. However, it computes the proper level of noise by iterating through $\mathcal{D}^u_{tr}$ before running actual optimization. SCRUB, which is based on knowledge distillation, requires several passes through the $\mathcal{D}^u_{tr}$ and hence does not show any progress after one epoch. In summary, contrastive unlearning is most efficient as it achieves unlearning by only requiring 60 batches to achieve unlearning.

Table~\ref{tab:sc_time} shows the elapsed time for each unlearning algorithm. Contrastive unlearning is the fastest among all baselines and across all models because it only requires running a single iteration over unlearning samples. The speed of UNSIR is the second fastest as it also runs for a single iteration, however, extra time has been consumed from computing adequate noise to perturb parameters.

\subsection{Results on Sample Unlearning}

\begin{table}[t]
\scriptsize
    \centering
    \setlength{\tabcolsep}{1pt}
    \begin{tabularx}{0.5\textwidth}{*{8}{>{\centering\arraybackslash}X}}
        \toprule
        \scalebox{0.9}{Model} & \scalebox{0.9}{Evaluation} & \scalebox{0.9}{Retrain} & \scalebox{0.9}{\textbf{Contrastive}} & \scalebox{0.9}{Finetune} & \scalebox{0.9}{Neggrad} & \scalebox{0.9}{Fisher} & \scalebox{0.9}{LCODEC} \\
        
        \midrule 
        \multirow{2}{*}{RN18}  & $\mathcal{D}^r_{ts}$ & 84.47 & 82.82 & 81.44 & 68.23 & 77.40 & 76.52\\
                                    & $\mathcal{D}^u_{tr}$ & 85.60 & 82.20 & 85.40 & 92.60 & 96.00 & 99.80\\
        \midrule
        \multirow{2}{*}{RN34}  & $\mathcal{D}^r_{ts}$ & 85.88 & 82.45 & 83.69 & 62.42 & 75.31 & 80.38\\
                                    & $\mathcal{D}^u_{tr}$ & 85.80 & 82.20 & 88.40 & 89.20 & 93.20 & 99.90\\
        \midrule
        \multirow{2}{*}{RN50}  & $\mathcal{D}^r_{ts}$ & 85.44 & 85.06 & 82.85 & 71.69 & 71.44 & 77.49\\
                                    & $\mathcal{D}^u_{tr}$ & 87.00 & 83.00 & 85.60 & 90.80 & 87.00 & 99.80\\
        \midrule
        \multirow{2}{*}{RN101} & $\mathcal{D}^r_{ts}$ & 86.00 & 86.24 & 75.38 & 76.92 & 82.15 & 78.03\\
                                    & $\mathcal{D}^u_{tr}$ & 85.69 & 84.20 & 77.60 & 96.00 & 97.80 & 99.80\\
                        
        \bottomrule
    \end{tabularx}
    \caption{Performance evaluation on sample unlearning on CIFAR-10.}
    \label{tab:rs_cifar10}
\end{table}
\begin{table}[t]
\scriptsize
    \centering
    
    \setlength{\tabcolsep}{1pt}
    \begin{tabularx}{0.5\textwidth}{*{8}{>{\centering\arraybackslash}X}}
        \toprule
        \scalebox{0.9}{Model} & \scalebox{0.9}{Evaluation} & \scalebox{0.9}{Retrain} & \scalebox{0.9}{\textbf{Contrastive}} & \scalebox{0.9}{Finetune} & \scalebox{0.9}{Neggrad} & \scalebox{0.9}{Fisher} & \scalebox{0.9}{LCODEC} \\
        
        \midrule 
        \multirow{2}{*}{RN18}  & $\mathcal{D}^r_{ts}$ & 95.08 & 91.16 & 90.77 & 79.96 & 88.06 & 93.89 \\
                                    & $\mathcal{D}^u_{tr}$ & 94.40 & 89.60 & 90.20 & 94.40 & 94.80 & 99.80 \\
        \midrule
        \multirow{2}{*}{RN34}  & $\mathcal{D}^r_{ts}$ & 95.49 & 92.43 & 91.74 & 73.93 & 93.00 & 94.62 \\
                                    & $\mathcal{D}^u_{tr}$ & 94.40 & 92.00 & 90.80 & 92.80 & 99.20 & 99.80 \\
        \midrule
        \multirow{2}{*}{RN50}  & $\mathcal{D}^r_{ts}$ & 95.93 & 93.23 & 91.37 & 89.22 & 91.16 & 93.70 \\
                                    & $\mathcal{D}^u_{tr}$ & 95.00 & 93.00 & 89.40 & 96.60 & 96.60 & 99.95 \\
        \midrule
        \multirow{2}{*}{RN101} & $\mathcal{D}^r_{ts}$ & 96.01 & 92.26 & 91.46 & 88.65 & 95.77 & 81.65 \\
                                    & $\mathcal{D}^u_{tr}$ & 93.80 & 90.40 & 91.00 & 97.60 & 99.80 & 91.60 \\
                        
        \bottomrule
    \end{tabularx}
    \caption{Performance evaluation on sample unlearning on SVHN.}
    \label{tab:rs_svhn}
\end{table}

\partitle{Model Performance and Unlearning Effectiveness}
Table~\ref{tab:rs_cifar10} shows accuracy on $\mathcal{D}^u_{tr}$ (unlearning samples) and $\mathcal{D}_{ts}$ (test data) on CIFAR-10 dataset. The retrain model shows the desired result as accuracy on $\mathcal{D}^u_{tr}$ (unlearning accuracy) and accuracy on $\mathcal{D}_{ts}$ (test accuracy) are similar. Contrastive unlearning exhibits the most similar performance with  the retrain model. Neggrad, Fisher, and LCODEC present higher unlearning accuracy compared to test accuracy, indicating that these models still have information that helps classify the unlearning samples. Finetune, while showing reasonable results, its performance varies a lot with model architecture and has significant model utility loss in some cases.

Table~\ref{tab:rs_svhn} presents test and unlearning accuracy on the SVHN dataset. LCODEC and Fisher show similar test accuracy with the retrain model on some models. However, their unlearning accuracy is very high, at almost 100\%, indicating a significant residual of the influence. 
Both Finetune and Neggrad show significant performance loss in test accuracy. Contrastive unlearning is more consistent in achieving similar unlearning accuracy as the retrain model with a relatively small performance loss in test accuracy.  

\begin{table}[t]
    \centering
    \scriptsize
    \setlength{\tabcolsep}{1pt}
    \begin{tabularx}{0.5\textwidth}{*{8}{>{\centering\arraybackslash}X}}
        \toprule
        \scalebox{0.9}{Model} & \scalebox{0.9}{Evaluation} & \scalebox{0.9}{Retrain} & \scalebox{0.9}{\textbf{Contrastive}} & \scalebox{0.9}{Finetune} & \scalebox{0.9}{Neggrad} & \scalebox{0.9}{Fisher} & \scalebox{0.9}{LCODEC} \\
        \midrule
        \multirow{2}{*}{RN18} & $\mathcal{D}^u_{tr}$ & 61.77  & 60.88 & 65.89 & 80.46 & 86.95 & 92.56 \\
			       & $\mathcal{D}^r_{tr}$ & 96.41  & 91.35          & 86.50 & 82.69 & 88.24 & 93.09 \\ \midrule 
        \multirow{2}{*}{RN34} & $\mathcal{D}^u_{tr}$ & 64.92  & 53.59 & 67.73 & 83.20 & 82.21 & 95.49 \\
			       & $\mathcal{D}^r_{tr}$ & 94.82  & 86.98          & 87.15 & 82.35 & 83.96 & 97.51 \\ \midrule 
        \multirow{2}{*}{RN50} & $\mathcal{D}^u_{tr}$ & 63.31  & 60.23 & 69.75 & 86.59 & 74.29 & 94.56 \\
			       & $\mathcal{D}^r_{tr}$ & 97.23  & 90.31          & 84.17 & 89.53 & 76.83 & 92.27 \\
        \midrule
        \multirow{2}{*}{RN101} & $\mathcal{D}^u_{tr}$ & 62.39  & 60.25 & 55.37 & 92.57 & 84.20 & 94.93 \\
			       & $\mathcal{D}^r_{tr}$ & 95.90  & 86.45 & 59.43          & 91.76 & 85.70 & 95.90 \\
        \bottomrule
    \end{tabularx}
    \caption{Member prediction rate on $\mathcal{D}^u_{tr}$ (unlearning samples) and $\mathcal{D}^r_{tr}$ (member-test samples) of MIA on CIFAR-10 dataset.}
    \label{tab:mia}
\end{table}

\partitle{Unlearning Effectiveness via MIA}
Table~\ref{tab:mia} shows the member prediction rate of the MIA on unlearning samples and test member samples against each unlearned model. 
An ideal attack model against the retrain model should have zero member prediction rate for unlearning samples and 100\% for member samples (since the unlearning samples are non-members). However, the attack model in our experiment shows around 60\% for unlearning samples which is a technical limitation of the attack model. The high rate on member samples does suggest that it has reasonable attack power in recognizing members. An unlearning algorithm is more effective if it exhibits  1) lower member prediction rate on unlearning samples, and 2) bigger difference in member prediction rate on unlearning samples and member samples.
For Neggrad, Fisher, and LCODEC, the member prediction rate for member samples and unlearning samples are similar, showing ineffective unlearning.
For finetune and contrastive unlearning, the member prediction rate for unlearning samples is lower than member samples. 
However, the difference is significantly bigger in contrastive unlearning, suggesting stronger discrimination between unlearning samples and member samples and more effective unlearning. 

\begin{table}
    \centering
    \scriptsize
    \setlength{\tabcolsep}{0.5pt}
    \begin{tabularx}{0.5\textwidth}{*{7}{>{\centering\arraybackslash}X}}
        \toprule
        Model & Retrain & \textbf{Contrastive} & Finetune & Neggrad & Fisher & LCODEC \\
        \midrule
        RN18 & 41.01 & \textbf{2.64} & 18.92 & 4.32 & 75.23 & 37.62 \\
        RN34 & 71.01 & 3.51 & 31.51 & \textbf{3.15} & 115.51 & 55.50 \\
        RN50 & 131.51 & \textbf{8.46} & 38.49 & 13.91 & 223.56 & 153.02 \\
        RN101 & 215.53 & \textbf{12.63} & 101.88 & 23.56 & 407.54 & 495.01 \\
        \bottomrule
    \end{tabularx}
    \caption{Processing time of each algorithm conducting sample unlearning on CIFAR-10 dataset (in minutes).}
    \label{tab:time}
\end{table}

\partitle{Efficiency}
Table~\ref{tab:time} shows the runtime of different algorithms. It clearly shows contrastive unlearning is overall the fastest. This is due to its fast convergence; it needs less than 20 iterations over the entire unlearning samples. While Neggrad also iterates only on unlearning samples, it requires more than 40 iterations to achieve unlearning effects. Finetune, Fisher, and LCODEC need longer runtime since they require iterating over the remaining samples. Fisher and LCODEC suffer excessive computation with larger models because their mathematical computation is proportional to model parameters and hardly parallelizable.

\section{Conclusion}
In this paper, we proposed a novel contrastive approach for machine unlearning. It achieves unlearning by re-configuring geometric properties of embedding space and contrasting unlearning samples and remaining samples. Through extensive experiments, we demonstrated that it outperforms state-of-the-art unlearning algorithms in model performance, unlearning effectiveness, and efficiency. 
In future work, we will examine the effectiveness of contrastive unlearning in different model architectures and different unlearning scenarios such as graph unlearning and correlated sequence unlearning.

\appendix





\clearpage
\bibliographystyle{named}
\bibliography{ijcai24}

\begin{thebibliography}{}

\bibitem[\protect\citeauthoryear{Arpit \bgroup \em et al.\egroup }{2017}]{arpit_closer_2017}
Devansh Arpit, Stanislaw Jastrzebski, Nicolas Ballas, David Krueger, Emmanuel Bengio, Maxinder~S. Kanwal, Tegan Maharaj, Asja Fischer, Aaron Courville, Yoshua Bengio, and Simon Lacoste-Julien.
\newblock A closer look at memorization in deep networks.
\newblock In {\em Proceedings of the 34th International Conference on Machine Learning}, pages 233--242. PMLR, 2017.
\newblock {ISSN}: 2640-3498.

\bibitem[\protect\citeauthoryear{Bourtoule \bgroup \em et al.\egroup }{2021}]{bourtoule_machine_2021}
Lucas Bourtoule, Varun Chandrasekaran, Christopher~A. Choquette-Choo, Hengrui Jia, Adelin Travers, Baiwu Zhang, David Lie, and Nicolas Papernot.
\newblock Machine unlearning.
\newblock In {\em 2021 {IEEE} Symposium on Security and Privacy ({SP})}, pages 141--159. {IEEE}, 2021.

\bibitem[\protect\citeauthoryear{Cao and Yang}{2015}]{cao_towards_2015}
Yinzhi Cao and Junfeng Yang.
\newblock Towards making systems forget with machine unlearning.
\newblock In {\em 2015 {IEEE} Symposium on Security and Privacy}, pages 463--480. {IEEE}, 2015.

\bibitem[\protect\citeauthoryear{Chen \bgroup \em et al.\egroup }{2020}]{chen_simple_2020}
Ting Chen, Simon Kornblith, Mohammad Norouzi, and Geoffrey Hinton.
\newblock A simple framework for contrastive learning of visual representations.
\newblock In {\em Proceedings of the 37th International Conference on Machine Learning}, pages 1597--1607. {PMLR}, 2020.
\newblock {ISSN}: 2640-3498.

\bibitem[\protect\citeauthoryear{Chen \bgroup \em et al.\egroup }{2023}]{chen_boundary_2023}
Min Chen, Weizhuo Gao, Gaoyang Liu, Kai Peng, and Chen Wang.
\newblock Boundary unlearning: Rapid forgetting of deep networks via shifting the decision boundary.
\newblock In {\em Proceedings of the IEEE/CVF Conference on Computer Vision and Pattern Recognition}, pages 7766--7775, 2023.

\bibitem[\protect\citeauthoryear{Cotogni \bgroup \em et al.\egroup }{2023}]{cotogni_duck_2023}
Marco Cotogni, Jacopo Bonato, Luigi Sabetta, Francesco Pelosin, and Alessandro Nicolosi.
\newblock {DUCK}: {Distance}-based {Unlearning} via {Centroid} {Kinematics}, December 2023.
\newblock arXiv:2312.02052 [cs].

\bibitem[\protect\citeauthoryear{Das and Chaudhuri}{2024}]{das_separability_2019}
Rudrajit Das and Subhasis Chaudhuri.
\newblock On the separability of classes with the cross-entropy loss function, 2024.

\bibitem[\protect\citeauthoryear{Golatkar \bgroup \em et al.\egroup }{2020}]{golatkar_eternal_2020}
Aditya Golatkar, Alessandro Achille, and Stefano Soatto.
\newblock Eternal sunshine of the spotless net: Selective forgetting in deep networks.
\newblock In {\em 2020 {IEEE}/{CVF} Conference on Computer Vision and Pattern Recognition ({CVPR})}, pages 9301--9309. {IEEE}, 2020.

\bibitem[\protect\citeauthoryear{Goodfellow \bgroup \em et al.\egroup }{2013}]{goodfellow_empirical_2013}
I.~Goodfellow, Mehdi Mirza, Xia Da, Aaron~C. Courville, and Yoshua Bengio.
\newblock An {Empirical} {Investigation} of {Catastrophic} {Forgeting} in {Gradient}-{Based} {Neural} {Networks}.
\newblock {\em CoRR}, December 2013.

\bibitem[\protect\citeauthoryear{Graf \bgroup \em et al.\egroup }{2021}]{graf_dissecting_2021}
Florian Graf, Christoph Hofer, Marc Niethammer, and Roland Kwitt.
\newblock Dissecting supervised contrastive learning.
\newblock In {\em Proceedings of the 38th International Conference on Machine Learning}, pages 3821--3830. {PMLR}, 2021.
\newblock {ISSN}: 2640-3498.

\bibitem[\protect\citeauthoryear{Guo \bgroup \em et al.\egroup }{2020}]{guo_certified_2020}
Chuan Guo, Tom Goldstein, Awni Hannun, and Laurens Van Der~Maaten.
\newblock Certified data removal from machine learning models.
\newblock In {\em Proceedings of the 37th International Conference on Machine Learning}, volume 119 of {\em {ICML}'20}, pages 3832--3842. {JMLR}.org, 2020.

\bibitem[\protect\citeauthoryear{Gupta \bgroup \em et al.\egroup }{2021}]{gupta_adaptive_2021}
Varun Gupta, Christopher Jung, Seth Neel, Aaron Roth, Saeed Sharifi-Malvajerdi, and Chris Waites.
\newblock Adaptive machine unlearning.
\newblock In {\em Advances in Neural Information Processing Systems}, volume~34, pages 16319--16330. Curran Associates, Inc., 2021.

\bibitem[\protect\citeauthoryear{He \bgroup \em et al.\egroup }{2016}]{he_deep_2016}
Kaiming He, Xiangyu Zhang, Shaoqing Ren, and Jian Sun.
\newblock Deep residual learning for image recognition.
\newblock In {\em 2016 {IEEE} Conference on Computer Vision and Pattern Recognition ({CVPR})}, pages 770--778. {IEEE}, 2016.

\bibitem[\protect\citeauthoryear{Jia \bgroup \em et al.\egroup }{2023}]{jia2023model}
Jinghan Jia, Jiancheng Liu, Parikshit Ram, Yuguang Yao, Gaowen Liu, Yang Liu, Pranay Sharma, and Sijia Liu.
\newblock Model sparsity can simplify machine unlearning.
\newblock In {\em Neural Information Processing Systems}, 2023.

\bibitem[\protect\citeauthoryear{Khosla \bgroup \em et al.\egroup }{2020}]{khosla_supervised_2020}
Prannay Khosla, Piotr Teterwak, Chen Wang, Aaron Sarna, Yonglong Tian, Phillip Isola, Aaron Maschinot, Ce~Liu, and Dilip Krishnan.
\newblock Supervised contrastive learning.
\newblock In {\em Advances in Neural Information Processing Systems}, volume~33, pages 18661--18673. Curran Associates, Inc., 2020.

\bibitem[\protect\citeauthoryear{Koch and Soll}{2023}]{koch_no_2023}
Korbinian Koch and Marcus Soll.
\newblock No matter how you slice it: Machine unlearning with {SISA} comes at the expense of minority classes.
\newblock In {\em 2023 {IEEE} Conference on Secure and Trustworthy Machine Learning ({SaTML})}, pages 622--637, 2023.

\bibitem[\protect\citeauthoryear{Kurmanji \bgroup \em et al.\egroup }{2023}]{kurmanji_towards_2023}
Meghdad Kurmanji, Peter Triantafillou, Jamie Hayes, and Eleni Triantafillou.
\newblock Towards unbounded machine unlearning, 2023.

\bibitem[\protect\citeauthoryear{Lin \bgroup \em et al.\egroup }{2023}]{lin_erm-ktp_2023}
Shen Lin, Xiaoyu Zhang, Chenyang Chen, Xiaofeng Chen, and Willy Susilo.
\newblock Erm-ktp: Knowledge-level machine unlearning via knowledge transfer.
\newblock In {\em Proceedings of the IEEE/CVF Conference on Computer Vision and Pattern Recognition}, pages 20147--20155, 2023.

\bibitem[\protect\citeauthoryear{Mantelero}{2024}]{mantelero_eu_2013}
Alessandro Mantelero.
\newblock The {EU} proposal for a general data protection regulation and the roots of the ‘right to be forgotten’.
\newblock {\em Computer Law \& Security Review}, 29(3):229--235, 2024.

\bibitem[\protect\citeauthoryear{Mehta \bgroup \em et al.\egroup }{2022}]{mehta_deep_2022}
Ronak Mehta, Sourav Pal, Vikas Singh, and Sathya~N Ravi.
\newblock Deep unlearning via randomized conditionally independent hessians.
\newblock In {\em Proceedings of the IEEE/CVF Conference on Computer Vision and Pattern Recognition}, pages 10422--10431, 2022.

\bibitem[\protect\citeauthoryear{Neel \bgroup \em et al.\egroup }{2024}]{neel_descent--delete_2021}
Seth Neel, Aaron Roth, and Saeed Sharifi-Malvajerdi.
\newblock Descent-to-delete: Gradient-based methods for machine unlearning.
\newblock In {\em Proceedings of the 32nd International Conference on Algorithmic Learning Theory}, pages 931--962. {PMLR}, 2024.
\newblock {ISSN}: 2640-3498.

\bibitem[\protect\citeauthoryear{Shokri \bgroup \em et al.\egroup }{2017}]{shokri_membership_2017}
Reza Shokri, Marco Stronati, Congzheng Song, and Vitaly Shmatikov.
\newblock Membership {Inference} {Attacks} {Against} {Machine} {Learning} {Models}.
\newblock In {\em 2017 {IEEE} {Symposium} on {Security} and {Privacy} ({SP})}, pages 3--18, May 2017.
\newblock ISSN: 2375-1207.

\bibitem[\protect\citeauthoryear{Shore and Johnson}{1981}]{shore_properties_1981}
J.~Shore and R.~Johnson.
\newblock Properties of cross-entropy minimization.
\newblock {\em IEEE Transactions on Information Theory}, 27(4):472--482, July 1981.
\newblock Conference Name: IEEE Transactions on Information Theory.

\bibitem[\protect\citeauthoryear{Tarun \bgroup \em et al.\egroup }{2023}]{tarun_fast_2023}
Ayush~K. Tarun, Vikram~S. Chundawat, Murari Mandal, and Mohan Kankanhalli.
\newblock Fast yet effective machine unlearning.
\newblock {\em {IEEE} Transactions on Neural Networks and Learning Systems}, pages 1--10, 2023.

\bibitem[\protect\citeauthoryear{Thudi \bgroup \em et al.\egroup }{2022}]{thudi_necessity_2022}
Anvith Thudi, Hengrui Jia, Ilia Shumailov, and Nicolas Papernot.
\newblock On the necessity of auditable algorithmic definitions for machine unlearning.
\newblock In {\em 31st {USENIX} Security Symposium ({USENIX} Security 22)}, pages 4007--4022, 2022.

\bibitem[\protect\citeauthoryear{Wu \bgroup \em et al.\egroup }{2023}]{wu_certified_2023}
Kun Wu, Jie Shen, Yue Ning, Ting Wang, and Wendy~Hui Wang.
\newblock Certified edge unlearning for graph neural networks.
\newblock In {\em Proceedings of the 29th {ACM} {SIGKDD} Conference on Knowledge Discovery and Data Mining}, pages 2606--2617. {ACM}, 2023.

\bibitem[\protect\citeauthoryear{Yan \bgroup \em et al.\egroup }{2022}]{yan_arcane_2022}
Haonan Yan, Xiaoguang Li, Ziyao Guo, Hui Li, Fenghua Li, and Xiaodong Lin.
\newblock Arcane: An efficient architecture for exact machine unlearning.
\newblock In {\em Proceedings of the Thirty-First International Joint Conference on Artificial Intelligence, IJCAI-22}, pages 4006--4013, 2022.

\end{thebibliography}

\end{document}